\documentclass{article}
\usepackage{ijcai20}

\usepackage{times}
\usepackage{soul}
\usepackage{url}
\usepackage[utf8]{inputenc}
\usepackage[small]{caption}
\usepackage{graphicx}
\usepackage{amsmath}
\usepackage{booktabs}
\usepackage{color}
\usepackage{amsthm}
\usepackage{amssymb}
\usepackage{algorithm}
\usepackage{algorithmic}
\usepackage{tabularx}
\usepackage{multirow}
\usepackage{array}
\usepackage{diagbox}
\title{Recurrent Dirichlet Belief Networks\\ for Interpretable Dynamic Relational Data Modelling}

\author{
Yaqiong Li$^1$\and
Xuhui Fan$^2$\footnote{Corresponding Author}\and
Ling Chen$^{1}$\and
Bin Li$^{3}$\and
Zheng Yu$^{4}$\And
Scott A. Sisson$^2$
\affiliations
$^1$Centre for Artificial Intelligence, University of Technology Sydney\\
$^2$School of Mathematics \& Statistics, University of New South Wales, Sydney\\
$^3$School of Computer Science, Fudan University\\
$^4$Department of Electrical and Computer Engineering, Univeristy of Alberta
\emails
yaqiong.li@student.uts.edu.au,
\{xuhui.fan,scott.sisson\}@unsw.edu.au,
ling.chen@uts.edu.au
}

\begin{document}

\maketitle

\begin{abstract}
The Dirichlet Belief Network~(DirBN) has been recently proposed as a promising approach in learning interpretable deep latent representations for objects. 
In this work, we leverage its interpretable modelling architecture and propose a deep dynamic probabilistic framework -- the Recurrent Dirichlet Belief Network~(Recurrent-DBN) -- to study interpretable hidden structures from dynamic relational data. The proposed Recurrent-DBN has the following merits: (1) it infers interpretable and organised hierarchical latent structures for objects within and across time steps; (2) it enables recurrent long-term temporal dependence modelling, which outperforms the one-order Markov descriptions in most of the dynamic probabilistic frameworks; (3) the computational cost scales to the number of positive links only. In addition, we develop a new inference strategy, which first upward-and-backward propagates latent counts and then downward-and-forward samples variables, to enable efficient Gibbs sampling for the Recurrent-DBN. We apply the Recurrent-DBN to dynamic relational data problems. The extensive experiment results on real-world data validate the advantages of the Recurrent-DBN over the state-of-the-art models in interpretable latent structure discovery and improved link prediction performance.

\end{abstract}

%%%%%%%%%%%%%%%%%%%%%%%%%%%%%%%%%%%%%%%%
%%%%%%%%%%%%%%%%%%%%%%%%%%%%%%%%%%%%%%%%
\section{Introduction}
%%%%%%%%%%%%%%%%%%%%%%%%%%%%%%%%%%%%%%%%
%%%%%%%%%%%%%%%%%%%%%%%%%%%%%%%%%%%%%%%%

Dynamic data is a common feature in many real-world applications,  including 
relational data analysis~\cite{mucha2010community,phan2015natural,yang2018dependent} for learning time-varying node interactions, and text modelling~\cite{guo2018deep,schein2019poisson} for exploring topic evolution. Modelling dynamic data has become a vibrant research topic,
with popular techniques ranging from non-Bayesian methods, such as Collaborative Filtering with Temporal Dynamics~(SVD++)~\cite{koren2009collaborative}, to Bayesian deep probabilistic frameworks such as Deep Poisson-Gamma Dynamical Systems~(DPGDS)~\cite{guo2018deep}. 
The main advantage of Bayesian deep probabilistic frameworks 
is the flexible model design and the strong modelling performance. However, most of these frameworks are static so that they cannot account for the evolution of relationships over time. It would be highly beneficial if the frameworks can be extended to the dynamic setting to enjoy the modelling advantages.

The Dirichlet Belief Network~(DirBN)~\cite{zhao2018dirichlet} has been proposed recently as a promising deep probabilistic framework for learning \emph{interpretable} deep latent structures. To date, the DirBN has mainly been used in two applications: (1) topic structure learning~\cite{zhao2018dirichlet}, where latent representations are used to model the word distribution for topics; and (2) relational models~\cite{fan2019scalable}, where latent representations model the nodes' membership distribution over communities. By constructing a deep architecture for latent distributions, the DirBN can model high-order dependence between topic-word distributions~(in topic models) and nodes' membership distributions~(in relational models).

In this work, we propose a Recurrent Dirichlet Belief Network~(Recurrent-DBN) to explore the complex latent structures in dynamic relational data. In addition to constructing an interpretable deep architecture for the data within individual time steps, we also study the temporal dependence in the dynamic relational data through (layer-to-layer) connections crossing consecutive time steps. Consequently, our Recurrent-DBN can describe long-term temporal dependence~(i.e., the dependence between the current variables and those in the previous several time steps), improving over the one-order Markov structures that usually describe the dependence between the current variables and those in the previous one time step only.  %in most Bayesian deep probabilistic models for dynamic data. 

For model inference, we further develop an efficient Gibbs sampling algorithm. Besides upward propagating latent counts as done by DirBN, we also introduce a backward step to propagate the counts from the current time step to the previous time steps. %We apply the Recurrent-DBN to the dynamic relational data setting.
Our experiments on real-world dynamic relational data show significant advantages of the Recurrent-DBN over the state-of-the-art models in tasks of interpretable latent structure discovery and link prediction. Similar to DirBN that can be considered as a self-contained module~\cite{zhao2018dirichlet}, our Recurrent-DBN could be flexibly adapted to account for dynamic data other than evolving relational data, such as time-varying counts and dynamic drifting text data.

We summarise this paper's main merits as follows:
\begin{description}
\item[Model] Recurrent structures are designed to model long term temporal dependence. Also, interpretable and organised latent structures are well explored; 
\item[Inference] An efficient Gibbs sampling method is devised that first upward-backward propagates latent counts and then downward-forward samples variable;
\item[Results] Significantly improved model performance in real-world dynamic relational models compared to the state-of-the-art, including better link prediction performance and enhanced interpretable latent structure visualisation.
\end{description}

%%%%%%%%%%%%%%%%%%%%%%%%%%%%%%%%%%%%%%%%
%%%%%%%%%%%%%%%%%%%%%%%%%%%%%%%%%%%%%%%%
\section{Background information of DirBN}
%%%%%%%%%%%%%%%%%%%%%%%%%%%%%%%%%%%%%%%%
%%%%%%%%%%%%%%%%%%%%%%%%%%%%%%%%%%%%%%%%

We first give a brief review of the DirBN model. In general, the DirBN constructs a {\it multi-stochastic} layered architecture to represent interpretable latent distributions for objects. We describe it within the relational data setting for illustrative purposes. Given a binary observed linkage matrix $\pmb{R}\in\{0, 1\}^{N\times N}$ for $N$ nodes, where $R_{ij}$ denotes whether node $i$ has a relation to node $j$, the DirBN constructs  an  $L$-layer and $K$-length community membership distribution ${\pmb{\pi}_i=}\{\pmb{\pi}_i^{(l)}\}_{l=1}^L$ for each node $i$. The generative process for the membership distributions $\{\pmb{\pi}_i^{(l)}\}_{l=1}^L$, as well as the observed matrix $\pmb{R}$, can be briefly described as:
{%For notational convenience, any parameters with index $0$ are set to zero.
\begin{enumerate}
 \item For $l=1, \ldots, L$
 \begin{enumerate}
 \item $\beta_{i'i}^{(l-1)}\sim\text{Gam}(c, \frac{1}{d}), \forall i,i'=1, \ldots, N$
 \item
 \begin{small}
 $\pmb{\pi}_i^{(l)}\sim\text{Dirichlet}(\pmb{\alpha}^{1 \times K}\pmb{1}(l=1)+ \sum_{i'}\beta_{i'i}^{(l-1)}\pmb{\pi}_{i'}^{(l-1)})$
 \end{small}
 \end{enumerate}
 \item $\pmb{X}_i\sim\text{Multinomial}(M;\pmb{\pi}_{i}^{(L)}), \forall i=1, \ldots, N$;
 \item $R_{ij}\sim \text{Bernoulli}\left(f(\pmb{X}_i, \pmb{X}_j)\right), \forall i,i'=1, \ldots, N$;
\end{enumerate}}

%\begin{enumerate}
% \item Generate $\pmb{\pi}_i^{(1)}\sim\text{Dirichlet}(\pmb{\alpha}^{1 \times K}), i=1, \ldots, N$
% \item For $l=2, \ldots, L$, generate
% \begin{enumerate}
% \item $\beta_{i'i}^{(l-1)}\sim\text{Gam}(c, \frac{1}{d}), i', i=1, \ldots, N$
% \item
% \begin{small}
% $\pmb{\pi}_i^{(l)}\sim\text{Dirichlet}(\sum_{i'}\beta_{i'i}^{(l-1)}\pmb{\pi}_{i'}^{(l-1)}), i=1, \ldots, N$
% \end{small}
% \end{enumerate}
% \item Generate %$\pmb{X}_i\sim\text{Multinomial}(M;\pmb{\pi}_{i}^{(L)}), {\small i=1, \ldots, N}$
% \item Generate $R_{ij}\sim \text{Bernoulli}\left(f(\pmb{X}_i, \pmb{X}_j)\right), {\small i,j=1, \ldots, N}$
%\end{enumerate}
where $\pmb{\alpha}^{1\times K}$ is a concentration parameter generating the membership distribution in the $1$st layer, $\beta_{i'i}^{(l-1)}$ represents the information propagation coefficient from node $i'$ to node $i$ in the $(l-1)$th layer, $c$ and $d$ are the hyper-parameters generating these propagation coefficients, $\pmb{X}_i$ is the latent count information for node $i$ and $M$ is the sum of these counts, and $f(\pmb{X}_i, \pmb{X}_j)$
% \textcolor{red}{(where $\pmb{\pi}_i=(\pmb{\pi}_i^{(1)},\ldots,\pmb{\pi}_i^{(L)})$)} 
represents the probabilistic function mapping a pair of membership distributions to a linkage probability. A larger value of $\beta_{i'i}^{(l-1)}$ indicates higher influence of $\pmb{\pi}_{i'}^{(l-1)}$ on the generation of  $\pmb{\pi}_{i}^{(l)}$. Therefore, $\beta_{i'i}^{(l-1)}$ is set to $0$ if node $i'$ is not connected to node $i$ in the observed data $\pmb{R}$.

It is difficult to directly implement efficient Gibbs sampling for the DirBN because the prior and posterior distributions of the membership distributions $\pmb{\pi}_i^{(l)}$ are not conjugate. To address this issue,  a strategy of first upward propagating latent counts and then downward sampling variables has been developed in \cite{zhao2018dirichlet}. Given the count information $\pmb{X}_i$ for node $i$, the DirBN upward propagates $\pmb{X}_i$ to all the nodes in the $(L-1)$th layer through a Chinese Restaurant Table~(CRT) distribution. Each node in the $(L-1)$th layer collects these propagated counts and uses their sum as its latent count $\pmb{X}_i^{(L-1)}$ in the $(L-1)$th layer. This procedure is repeated until the counts have been assigned to all layers. Thus, conjugate constructions can be created for each variable and thereby used to construct efficient Gibbs samplers. 

\section{Recurrent-DBN for dynamic relational data modeling}
%%%%%%%%%%%%%%%%%%%%%%%%%%%%%%%%%%%%%%%%
%%%%%%%%%%%%%%%%%%%%%%%%%%%%%%%%%%%%%%%%

To handle dynamic relational data, we attach an index $t$ to variables to denote the corresponding time step. Thus, the observed dynamic relational data can be described as $\pmb{R}\in\{0, 1\}^{N\times N\times T}$ for $N$ nodes at $T$ time steps, where $R_{ij, t}$ denotes whether node $i$ has relation to node $j$ at the $t$th time step. Each matrix $\{\pmb{R}_{-, t}\}_t\in\{0, 1\}^{N\times N}$ can be either asymmetric~(directional) or symmetric~(non-directional) and we do not consider self-linkages $\{R_{ii,t}\}_{i,t}$. 
%%%%%%%%%%%%%%%%%%%%%%%%%%%%%%%%%%%%%%%%
%%%%%%%%%%%%%%%%%%%%%%%%%%%%%%%%%%%%%%%%
\subsection{Recurrent-DBN for latent structure generation}
%%%%%%%%%%%%%%%%%%%%%%%%%%%%%%%%%%%%%%%%
%%%%%%%%%%%%%%%%%%%%%%%%%%%%%%%%%%%%%%%%
\label{sec:generative}

\begin{figure}[t]
\centering
\includegraphics[width = 0.48\textwidth]{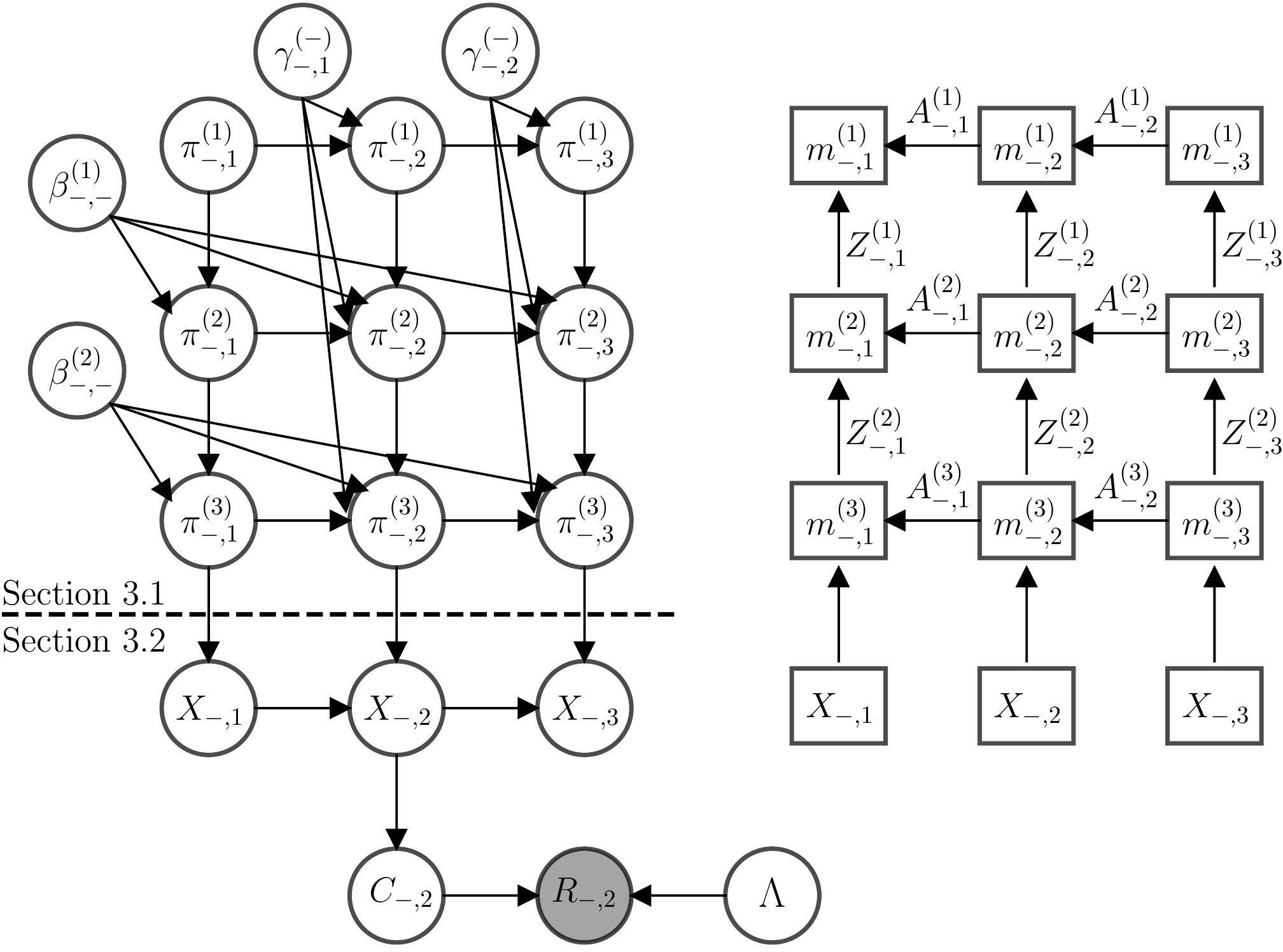}
\caption{Left: a brief graphical model of Recurrent-DBN with $3$-hidden-layers for a dynamic relational data with $3$ time steps~(Sections~\ref{sec:generative}\&\ref{sec:network_modelling}), where shaded nodes represent observed data. Hyper-parameters are ignored for concise presentation. Right: the upward-backward propagation of counts $\pmb{X}$ to each hidden layers in each {inference iteration} (Section \ref{sec:propagation}), where $\pmb{m}_{-,-}^{(-)}$ represents the latent counts attached to nodes at each layer and time step, $\pmb{Z}_{-,-}^{(-)}$ refers to the layer-wise propagated counts and $\pmb{A}_{-,-}^{(-)}$ is the propagated counts between consecutive time steps. } 
\label{fig:graphical_model}
\end{figure}

In the Recurrent-DBN, we assume the time-dependent membership distribution of a node $i$ in the $l$-th layer at time step $t$, $\pmb{\pi}_{i,t}^{(l)}$,  follows a Dirichlet distribution.
%each node can be characterized by a time-and-layer dependent latent representation $\pmb{\pi}_{i,t}^{(l)}$, where $\pmb{\pi}_{i,t}^{(l)}$ follow a Dirichlet distribution and represents node $i$'s latent distribution in the $l$-th layer at time step $t$. %In addition to modelling the time dependence, the deep hidden layers architecture enables us to induce the high-order node dependence. Let $\pmb{\pi}_{i,t}^{(l)}$ follow a Dirichlet distribution with $K$ categories, each $\pi_{i,k,t}^{(l)}$ can be considered as how strongly node $i$ associates with the $k$-th category in the $l$-th layer at time $t$. These categories could be considered as the unique word or communities/groups in topic structure learning or relational data modelling respectively.
Its generative process can be described as below, with the propagation of %latent distribution
$\pmb{\pi}_{i,t}^{(l)}$ illustrated in Fig.~\ref{fig:graphical_model}~(Left).
~For notation convenience, any parameters with index $0$ are set to zero. It is noted that we have already used the observed data into the generative process.
%through the whole text, all related variables are set as $0$ with situations of ${l-1}=0$ or ${t-1}=0$). 
\begin{enumerate}
 \item  For $t=1, \ldots, T, l=1, \ldots, L$ 
  \begin{enumerate}
 \item For $i',i=1, \ldots, N$
  \begin{enumerate}
  \item $\beta_{i'i,t}^{(l-1)}\left\{\begin{array}{ll}
  =0, & \text{ if }R_{i'i,t}=0; \\
  \sim\text{Gam}(c_c^{(l)}, \frac{1}{d_c}), & \text{ if }i'=i; \\
  \sim\text{Gam}(c_u^{(l)}, \frac{1}{d_c}), & \text{ if }i'\neq i .
  \end{array}\right.$
  \item $\gamma_{i'i,t-1}^{(l)}\left\{\begin{array}{ll}
  =0, & \text{ if }R_{i'i,t-1}=0; \\
  \sim\text{Gam}(c_c^{(l)}, \frac{1}{d_c}), & \text{ if }i'=i; \\
  \sim\text{Gam}(c_u^{(l)}, \frac{1}{d_c}), & \text{ if }i'\neq i. 
  \end{array}\right.$
     \end{enumerate}
 \item For $i=1, \ldots, N$
  \begin{enumerate}
   \item Calculate concentration parameter $\pmb{\psi}_{i,t}^{(l)}$:
     \begin{small}
     \begin{align}
      {\pmb{\psi}_{i,t}^{(l)}=\sum_{i'}\beta_{i'i, t}^{(l-1)}\pmb{\pi}_{i',t}^{(l-1)}+\sum_{i'}\gamma_{i'i, t-1}^{(l)}\pmb{\pi}_{i', t-1}^{(l)}}. \label{eq_pis}
      \end{align}
      \end{small}
   \item $\pmb{\pi}_{i,t}^{(l)}\sim\text{Dirichlet}(\pmb{\alpha}^{1 \times K}\pmb{1}(t=1,l=1)+ \pmb{\psi}_{i,t}^{(l)})$.
     \end{enumerate}
     \end{enumerate}
\end{enumerate}
Here, $\beta_{i'i,t}^{(l-1)}\in\mathbb{R}^+$ is the information propagation coefficient from node $i'$ in the $(l-1)$-th layer to node $i$ in the $l$-th layer at the same time $t$, $\gamma_{i'i,t-1}^{(l)}\in\mathbb{R}^+$ is the information propagation coefficient from node $i'$ at time $t-1$ to node $i$ at time $t$ in the same layer $l$,  $c_c^{(-)}, c_u^{(-)}, d_c$ are the corresponding hyper-parameters and $\pmb{\alpha}^{1 \times K}$ is the concentration parameter for the membership distribution in the first layer. The larger the value of these coefficients, the stronger the connections between the two corresponding latent representations~(i.e., $\pmb{\pi}_{i',t}^{(l-1)}$ and $\pmb{\pi}_{i, t}^{(l)}$, $\pmb{\pi}_{i',t-1}^{(l)}$ and $\pmb{\pi}_{i, t}^{(l)}$).

We restrict the two nodes to have information propagated only if they are observed with positive relationship~(step~(a).i and~(a).ii). This can reduce the computational cost of calculating $\pmb{\beta}_{-, t}^{(l)}, \pmb{\gamma}_{-, t}^{(l)}$ from  $\mathcal{O}(N^2)$ to the scale of the number of positive relationships. Also, it encourages connected nodes to have more similar membership distributions and larger dependencies between each other.

The concentration parameter $\pmb{\psi}_{i,t}^{(l)}$ for generating $\pmb{\pi}_{i,t}^{(l)}$ comprises two parts: the information propagated from all other nodes' latent representations in the $(l-1)$-th layer at time $t$, $\sum_{i'}\beta_{i'i, t}^{(l-1)}\pmb{\pi}_{i',t}^{(l-1)}$, and those in the $l$-th layer at time $(t-1)$, $\sum_{i'}\gamma_{i'i, t-1}^{(l)}\pmb{\pi}_{i', t-1}^{(l)}$. In other words, $\pmb{\psi}_{i,t}^{(l)}$ is a linear sum of all the previous-layers' information at the same time step and all the previous-time steps' information in the same layer. When the coefficients $\pmb{\beta}$ dominate over $\pmb{\gamma}$, the hierarchical structure plays a more important role. Otherwise, the temporal dependence has higher influence. 
%%%%%%%%%%%%%%%%%%%%%%%%%%%%%%%%%%%%%%%%
%%%%%%%%%%%%%%%%%%%%%%%%%%%%%%%%%%%%%%%%
\subsection{Application to dynamic relational data}
%%%%%%%%%%%%%%%%%%%%%%%%%%%%%%%%%%%%%%%%
%%%%%%%%%%%%%%%%%%%%%%%%%%%%%%%%%%%%%%%%
\label{sec:network_modelling}

After generating the membership distributions $\{\pmb{\pi}_{i,t}^{(l)}\}$, we use the Bernoulli-Poisson link function~\cite{dunson2005bayesian,zhou2015infinite,fan2019scalable} to generate the relational data at each time step:
\begin{enumerate}
    \item $\Lambda_{k_1k_2}\sim\text{Gamma}(\lambda_1, \lambda_0),\forall k_1,k_2$
   \item $M_{i,t}\sim\text{Poisson}(M), \forall i, t$
   \item $\pmb{X}_{i,t}\sim\text{Multinomial}(M_{i,t};\pmb{\pi}_{i,t}^{(L)}), \forall i, t$;
 \item For $t=1, \ldots, T,i,j=1, \ldots, N$,
   \begin{enumerate}
    \item $C_{ij,k_1k_2,t}\sim\text{Poisson}(X_{i,k_1,t}\Lambda_{k_1k_2}X_{j,k_2,t}), \forall k_1,k_2$
    \item $R_{ij,t}={\pmb{1}(\sum_{k_1,k_2} C_{ij,k_1k_2,t}>0)}$,
  \end{enumerate}
 \end{enumerate}
where $\Lambda_{k_1k_2}$ is a community compatibility parameter such that a larger value of $\Lambda_{k_1k_2}$ indicates a larger possibility of generating the links between communities $k_1$ and $k_2$, $\lambda_1, \lambda_0, M$ are hyper-parameters, $M_{i,t}$ is a scaling parameter for generating the related counting information for node $i$ at time $t$, and $C_{ij,k_1k_2,t}$ is a community-to-community latent integer for linkage $R_{ij}$ at time $t$.

%The $\{\pmb{X}_{i,t}\}_{i,t}$ here are $K$-length counting vectors obtaining 
Through the Multinomial distributions with $\pmb{\pi}_{i,t}^{(L)}$ as event probabilities, $\pmb{X}_{i,t}$ can be regarded as an estimator of $\pmb{\pi}_{i,t}^{(L)}$. Since the sum $M_i\sim{\text{Poisson}(M)}$, according to the Poisson-Multinomial equivalence, each $X_{i,k,t}$ is equivalently distributed as $X_{i,k,t}\sim{\text{Poisson}(M{\pi}_{i,k,t}^{(L)})}$. Therefore, both the prior distribution for generating $X_{i,k,t}$ and the likelihood based on $X_{i,k,t}$ are Poisson distributions. We may form feasible categorical distribution on its posterior inference.  This trick is inspired by the recent advances in data augmentation and marginalisation
techniques~\cite{fan2019scalable}, which allows us to implement posterior sampling for $X_{i,k,t}$ efficiently. 

The counts $\pmb{X}_{i,t}$ lead to the generation of the $K \times K$ integer matrix $\pmb{C}_{ij,t}$. Based on the Bernoulli-Poisson link function~\cite{dunson2005bayesian,zhou2015infinite}, the observed $R_{ij,t}$ is mapped to the latent Poisson count random variable matrix $\pmb{C}_{ij,t}$. It is shown in~\cite{fan2019scalable} that $\{C_{ij,k_1k_2,t}\}_{k_1,k_2}=0$ if $R_{ij,t}=0$. That is, only the non-zero links are involved during the inference for $\pmb{C}_{ij,k_1k_2,t}$, which largely reduces the computational complexity, especially for large and sparse dynamic relational data.

\textbf{Recurrent structure.} Before describing the inference of Recurrent-DBN, we discuss the characteristic of the recurrent structure of our model. Instead of using the one-order Markov property to describe the temporal dependence~(assuming the state at time $t$ depends on the states at time $t-1$ only), which is adopted by most probabilistic dynamic models, the deep structure of the Recurrent-DBN allows the latent variables at time $t$ depend on those at time steps from $t-1$ to $t-L$. For example, by using the law of total expectations, we can have the expectation of the latent count $\pmb{X}_{-,t}$ in a $2$-layered Recurrent-DBN as~(We use the notation $-$ to denote a related parameter or variable hereafter):
%Rather than constructing the time-dependence at the time step $t$ purely on random variable the previous $t-1$ time step, $t-1$ in most conventional dynamic models, the deep structure of Recurrent-DBN {allows the correlation of} latent distributions {be considered based} on all the existing layers {and time steps}. By using the law of total expectations, for a $2$-layered Recurrent-DBN, we have:
{\begin{align} \label{recurrent_eq}
 \mathbb{E}\left[\pmb{X}_{-, t}|-\right]= & \pmb{\beta}^{(L-1)}_{-,t-1}\pmb{\gamma}^{(L-1)}_{-, t-1}\pmb{\pi}_{-,t-1}^{(L)}\nonumber \\
& +\pmb{\beta}^{(L-1)}_{-,t-1}\pmb{\beta}^{(L-2)}_{-,t-1}\pmb{\gamma}^{(L-2)}_{-,t-1}\pmb{\gamma}^{(L-2)}_{-,t-2}\pmb{\pi}_{-,t-2}^{(L-1)}.
\end{align}}
In Eq.~(\ref{recurrent_eq}), $\mathbb{E}\left[\pmb{X}_{-, t}|-\right]$ depends on both $\pmb{\pi}_{-,t-1}$ and $\pmb{\pi}_{-,t-2}$. This format can be extended straightforwardly to $L$-layers  and involve more previous membership distributions. Such recurrent structures allow us to summarise and abstract those random variables, capturing both the hierarchical latent structures and the dynamic dependencies. 

%%%%%%%%%%%%%%%%%%%%%%%%%%%%%%%%%%%%%%%%
%%%%%%%%%%%%%%%%%%%%%%%%%%%%%%%%%%%%%%%%
\subsection{Inference}
%%%%%%%%%%%%%%%%%%%%%%%%%%%%%%%%%%%%%%%%
%%%%%%%%%%%%%%%%%%%%%%%%%%%%%%%%%%%%%%%%
\label{sec:propagation}

%The related posterior inference involves the variables of $\{\pmb{\pi}_{i, t}^{(l)}\}_{i,t, l}, \{\beta_{i'i,t}^{(l)}, \gamma_{i'i,t}^{(l)}\}_{i',i, t, l}, \pmb{\Lambda}, \{\pmb{X}_{i,t}\}_{i,t}, \{\pmb{C}_{ij, t}\}_{i,j,t}$. 
The joint distribution of the latent variables is expressed as:
{\small \begin{align}
& P(\pmb{\pi}, \pmb{\beta}, \pmb{\gamma}, \pmb{X}, \pmb{C}, \pmb{\Lambda},\pmb{R}|-)=\prod_{i',i,t}P(R_{i',i,t}|\pmb{C})P(\pmb{\Lambda}|\lambda_1,\lambda_0)\nonumber \\
&\cdot \prod_{i,l, t}\left[P(\pmb{\pi}_{i,t}^{(l)}|\pmb{\pi}_{-,t-1}^{(l)}, \pmb{\pi}_{-,t}^{(l-1)}, \pmb{\beta}_{- i, t}^{(l)}, \pmb{\gamma}_{- i, t}^{(l)})\cdot P({\beta}_{- i, t}^{(l)}|-)P({\gamma}_{- i, t}^{(l)}|-)\right]\nonumber \\
&\cdot \prod_{i,t}\left[P(X_{i,t}|\pmb{\pi}_{i,t}, M)\prod_{i',k_1,k_2}P(C_{i'i,k_1k_2,t}|X_{i,t}, X_{j,t}, \Lambda_{k_1k_2})\right]\nonumber 
\end{align}}
{While the DirBN only has upward-propagation for the latent counts and downward-sampling for the latent variables, for the Recurrent-DBN we develop an upward-backward propagation and forward-downward Gibbs sampling algorithm for count propagation and latent variable sampling.
Posterior simulation for the Recurrent-DBN involves two key steps in each sampling iteration: (1) propagating the counts $\pmb{X}$ upward and backward to the upper layers and previous time steps via a latent count variable $\pmb{m}$; (2) forward and downward sampling $\pmb{\pi}, \pmb{\beta}, \pmb{\gamma}$ given the propagated latent counts $\pmb{m}$.} Full updates for the other variables are similar to those in~\cite{fan2019scalable}. 

%{While DirBN has only the upward propagation for latent counts and the downward sampling of latent variables, we develop an upward-backward count propagation and forward-downward Gibbs sampling of latent variables in Recurrent-DBN.  %due to its deep hierarchical structure.
%The inference of Recurrent-DBN involves two key steps in each iteration: (1) propagating the counts $\pmb{X}$ upward and backward to the upper layers and previous time steps via an latent count variable $\pmb{m}$; (2) forward and downward sampling $\pmb{\pi}, \pmb{\beta}, \pmb{\gamma}$ given the propagated latent count $\pmb{m}$.} The inference for the other variables is similar to~\cite{fan2019scalable} and we omit the  details for space constraints. 

\subsubsection{Upward-Backward Propagating the Latent Counts}
Figure~\ref{fig:graphical_model} (right) illustrates the upward-backward propagation of counts $\pmb{X}$ to the latent count variable $\pmb{m}$ at each hidden layers. Generally speaking, for $ i,i'=1, \ldots, N, l=1, \ldots, L,t=1, \ldots, T, k=1, \ldots, K$, the latent variable $\pmb{\psi}$ is generated as Eq.~(\ref{eq_pis}). $m_{i,k,t}^{(l)}$ refers to the latent counts for the node $i$ in layer $l$ at time $t$ for the $k$-th community. By integrating the $m_{i,k,t}^{(l)}$, the likelihood term of $\pmb{\psi}_{i,t}^{(l)}$ can be calculated as:
\begin{align}
 \mathcal{L}(\pmb{\psi}_{i,t}^{(l)})\propto \frac{\Gamma(\sum_k {\psi}_{i,k,t}^{(l)})}{\Gamma(\sum_k {\psi}_{i,k,t}^{(l)}+\sum_k m_{i,k,t}^{(l)})}\prod_k \frac{\Gamma({\psi}_{i,k,t}^{(l)}+m_{i,k,t}^{(l)})}{\Gamma({\psi}_{i,k,t}^{(l)})} \nonumber 
\end{align}
where $\Gamma(-)$ is a Gamma function. 

By introducing the auxiliary variables $q_{i,t}^{(l)}$ and $y_{i,k,t}^{(l)}$, the likelihood term of $\pmb{\psi}_{i,t}^{(l)}$ can be further augmented as:
\begin{align}
 \mathcal{L}(\pmb{\psi}_{i,t}^{(l)}, q_{i,t}^{(l)}, y_{i,k,t}^{(l)})\propto \prod_{k=1}^K \left(q_{i,t}^{(l)}\right)^{\psi_{i,k,t}^{(l)}}
 \left(\psi_{i,k,t}^{(l)}\right)^{y_{i,k,t}^{(l)}}\nonumber
\end{align}
where the $q_{i, t}^{(l)}$ and $y_{i,k,t}^{(l)}$ can be generated as:
{\small \begin{align}
 y_{i,k,t}^{(l)}\sim\text{CRT}(m_{i,k,t}^{(l)}, \psi_{i,k,t}^{(l)}),q_{i, t}^{(l)}\sim\text{Beta}(\sum_{k}\psi_{i,k,t}^{(l)}, \sum_{k}m_{i,k,t}^{(l)})\nonumber
 \end{align}}
Consequently, $y_{i,k,t}^{(l)}$ can be considered as the `derived latent counts' for node $i$ derived from the latent counts $m_{i,k,t}^{(l)}$. Each $y_{i,k,t}^{(l)}$ can then be upward and backward distributed based on the probabilities of $\psi_{i,k,t}^{(l)}$ as follows:
\begin{align}
& (Z_{i1,k,t}^{(l-1)}, \ldots, Z_{iN,k,t}^{(l-1)}, A_{i1, t-1,k}^{(l)}, \ldots, A_{i1,t-1,k}^{(l)})\nonumber \\ \sim
&\text{Multinomial}(y_{i,k,t}^{(l)};\frac{\pmb{\beta}_{-i, t}^{(l-1)}{\pmb{\pi}}_{-,k,t}^{(l-1)}}{\psi_{i,k,t}^{(l)}}, \frac{\pmb{\gamma}_{-i, t-1}^{(l)}{\pmb{\pi}}_{-,t-1,k}^{(l)}}{\psi_{i,k,t}^{(l)}}) \label{eq_Z_A}
\end{align}
Here, the $y_{i,k,t}^{(l)}$ is divided into two parts: one is delivered to each $i'$ at time $t$ of layer $l-1$ ($(Z_{i1,k,t}^{(l-1)}, \ldots, Z_{iN,k,t}^{(l-1)})$), and the other to each $i'$ at time $t-1$ of layer $l$ ($A_{i1, t-1,k}^{(l)}, \ldots, A_{i1,t-1,k}^{(l)})$). We denote them as $\pmb{Z}_{i-,k,t}^{(l-1)}$ and $\pmb{A}_{i-,t-1,k}^{(l)}$ respectively.
The latent counts of lower layers and previous time steps can thus be calculated respectively as:
\begin{align}
 &m_{i,k,t}^{(l-1)}=\sum_{i'}Z_{i'i,k,t}^{(l-1)}+\sum_{i'}A_{i'i,k,t}^{(l-1)} \nonumber \\
 &m_{i,t-1,k}^{(l)}=\sum_{i'}Z_{i'i,t-1,k}^{(l)}+\sum_{i'}A_{i'i,t-1,k}^{(l)} \label{eq_X_l}
\end{align}
 Let $\pmb{m}_{i, T}^{(L)}=\pmb{X}_{i,T}$, 
for $t=T-1, \ldots, 2,i,i'=1, \ldots, N$, the specification in terms of layer $L$ is as follows, 
\begin{align}
m_{i,t-1,k}^{(L)}=X_{i,t-1,k}+\sum_{i'}A_{i'i,t-1,k}^{(L)} \label{eq_X_L}
\end{align}
%All the related variable is set as $0$ in any situation associated with layer $l-1=0$ or time $t-1=0$. 
%Thus, by introducing a few auxiliary variables and implementing several data augmentation techniques, the propagation allows efficient layer and time-wise Gibbs sampling for all the latent variables. The whole process is illustrated in Figure~(\ref{fig:graphical_model}(b)).

To summarize, upward and backward propagation derives $\pmb{y}_{i,t}^{(l)}$ from the latent counts $\pmb{m}_{i,t}^{(l)}$. Then,  $\pmb{y}_{i,t}^{(l)}$ is distributed to all $i'$ at time $t$ of layer $l-1$ and time $t-1$ of layer $l$ respectively as $\pmb{Z}_{i-,k,t}^{(l-1)}$ and $\pmb{A}_{i-,t-1,k}^{(l)}$. Lastly, $\pmb{Z}_{-i,k,t}^{(l-1)}$ and $\pmb{A}_{-i,t-1,k}^{(l)}$ contribute to the generation of $\pmb{m}_{i,t}^{(l-1)}$ and $\pmb{m}_{i,t-1}^{(l)}$ respectively. By repeating this process through layers and crossing time steps, we propagate the $\pmb{X}$ to the $\pmb{m}^{(l)}$ upward and backward sequentially.

\subsubsection{Forward-Downward Sampling Latent Variables}
The generated $\pmb{\psi}, \pmb{q}, \pmb{m}^{(l)}, (\pmb{Z},\pmb{A})$ can enable to form closed Gibbs sampling algorithm for the following variables:
\paragraph{Sampling $\{\pmb{\pi}_{i, t}^{(l)}\}_{i,t,l}$}
After obtaining the latent counts $\pmb{m}_{i,t}^{(l)}$ for each layer and each time step, the posterior inference of $\pmb{\pi}_{i,t}^{(l)}$ can be proceeded as:
\begin{align} %\label{eq_pi}
\pmb{\pi}_{i,t}^{(l)}\sim\text{Dirichlet}(\pmb{\psi}_{i,t}^{(l)}+\pmb{m}_{i,t}^{(l)})\nonumber
%, \forall i, l, t
\end{align}

\paragraph{Sampling $\{{\beta}_{i'i,t}^{(l)},{\gamma}_{i'i,t}^{(l)} \}_{i', i,l,t}$} 

The likelihood term of ${\beta}_{i'i, t}^{(l)}$ can be represented as:
\begin{align} 
\mathcal{L}(\beta_{i'i, t}^{(l)})\propto e^{\log q_{i, t}^{(l)}\beta_{i'i, t}^{(l)}}\left(\beta_{i'i, t}^{(l)}\right)^{\sum_k Z_{i'i,k,t}^{(l)}} \nonumber
\end{align}

The prior of $\beta_{i'i, t}^{(l)}, \beta_{i'i, t}^{(l)}$ is $\text{Gam}(\gamma^{(l)}_{i}, \frac{1}{c^{(l)}})$. Their posterior distribution is
\begin{align} %\label{eq_beta_ii}
\beta_{i'i, t}^{(l)}\sim &\text{Gam}(\gamma_i^{(l)}+\sum_{k}Z_{i'i,k,t}^{(l)}, \frac{1}{c^{(l)}-\log q_{i',t}^{(l)}})\nonumber\\
\gamma_{i'i, t}^{(l)}\sim &\text{Gam}(\gamma_i^{(l)}+\sum_{k}A_{i'i,k,t}^{(l)}, \frac{1}{c^{(l)}-\log q_{i',t}^{(l)}}) \nonumber
\end{align}

%\paragraph{Sampling $\{{\gamma}_{i'i,t}^{(l)}\}_{i', i,t, l}$} The sampling of ${\gamma}_{i'i, t}^{(l)}$ follows a similar implementation of ${\beta}_{i'i, t}^{(l)}$. Its likelihood term can be represented as:
%\begin{align} 
%\mathcal{L}(\gamma_{i'i, t}^{(l)})\propto e^{\log q_{i, t}^{(l)}\gamma_{i'i, t}^{(l)}}\left(\gamma_{i'i, t}^{(l)}\right)^{\sum_k A_{i'i,k,t}^{(l)}} \nonumber
%\end{align}

%The prior of $\gamma_{i'i, t}^{(l)}$ is $\text{Gam}(\gamma^{(l)}_{i}, \frac{1}{c^{(l)}})$, its posterior distribution is
%\begin{align} %\label{eq_gamma_ii}
%\gamma_{i'i, t}^{(l)}\sim \text{Gam}(\gamma_i^{(l)}+\sum_{k}A_{i'i,k,t}^{(l)}, \frac{1}{c^{(l)}-\log q_{i,t}^{(l)}}) \nonumber
%\end{align}

\begin{table*}[t]
\centering 
\small
\caption{Links prediction performance comparison. Note:* represents a dynamic model.}
\begin{tabular}{l r r r r r}
  \toprule
   \multicolumn{6}{c}{AUC (mean and standard deviation)} \\
  Model & Coleman & Mining reality & Hypertext & Infectious & Student net \\
  \midrule
  MMSB & $0.875\pm0.013$ & $0.883\pm0.009$ & $0.869\pm0.008$ & $0.969\pm0.004$ & $0.916\pm0.001$ \\
  $\text{T-MBM}^*$ & $0.886\pm0.012$ & $0.863\pm0.005$ & $0.797\pm0.009$ & $0.833\pm0.018$ & $0.886\pm0.015$ \\
  $\text{fcMMSB}^*$ & $0.909\pm0.005$ & $0.932\pm0.006$ & $0.909\pm0.005$ & $0.980\pm0.002$ & $0.958\pm0.003$ \\
  $\text{BPTF}^*$ & $0.907\pm0.003$ & $0.923\pm0.004$ & $0.871\pm0.006$ & $0.845\pm0.001$ & $0.905\pm0.011$ \\ 
  $\text{DRGPM}^*$ & $\cdots$ & $0.935\pm0.013$ & $0.906\pm0.002$ & $0.988\pm0.001$ & $0.825\pm0.004$ \\ 
  CN & $0.871\pm0.008$ & $0.863\pm0.014$ & $0.786\pm0.016$ & $0.889\pm0.004$ & $0.849\pm0.019$ \\
  $\text{SVD++}^*$ & $\cdots$ & $0.843\pm0.016$ & $0.725\pm0.014$ & $0.617\pm0.001$ & $\cdots$ \\
  MNE & $0.893\pm0.004$ & $0.823\pm0.004$ & $0.869\pm0.008$ & $0.898\pm0.007$ & $0.942\pm0.001$ \\ 
  DeepWalk & $0.916\pm0.008$ & $0.762\pm0.014$ & $0.826\pm0.015$ & $0.915\pm0.010$ & $0.915\pm0.008$ \\ 
  \midrule
  Recurrent-DBN,K=30 & $\textbf{0.919}\pm0.012$ & $\textbf{0.969}\pm0.000$ & $\textbf{0.944}\pm0.004$ & $\textbf{0.995}\pm0.000$ & $\textbf{0.976}\pm0.002$ \\ 
  Recurrent-DBN,K=20 & $0.909\pm0.019$ & $\textbf{0.965}\pm0.001$ & $\textbf{0.932}\pm0.003$ & $\textbf{0.995}\pm0.000$ & $\textbf{0.971}\pm0.002$ \\ 
  %Recurrent-DBN,KK=15 & $\textbf{0.909}\pm0.015$ & $\cdots$ & $\textbf{0.964}\pm0.001$ & $\textbf{0.930}\pm0.003$ & $\textbf{0.993}\pm0.001$ \\
  Recurrent-DBN,K=10 & $0.899\pm0.011$ & $\textbf{0.961}\pm0.002$ & $\textbf{0.926}\pm0.002$ & $\textbf{0.989}\pm0.000$ & $\textbf{0.964}\pm0.010$ \\ 
  \bottomrule
   \toprule
   \multicolumn{6}{c}{Precision (mean and standard deviation)} \\
  Model & Coleman & Mining reality & Hypertext & Infectious & Student net \\
  \midrule
  MMSB & $0.289\pm0.025$ & $0.126\pm0.009$ & $0.121\pm0.019$ & $0.233\pm0.065$ & $0.238\pm0.017$ \\
  $\text{T-MBM}^*$ & $0.199\pm0.015$ & $0.443\pm0.016$ & $0.142\pm0.010$ & $0.393\pm0.065$ & $0.168\pm0.007$ \\ 
  $\text{fcMMSB}^*$ & $0.344\pm0.017$ & $0.835\pm0.017$ & $0.505\pm0.012$ & $0.326\pm0.011$ & $0.304\pm0.007$ \\
  $\text{BPTF}^*$ & $0.385\pm0.057$ & $0.701\pm0.013$ & $0.297\pm0.010$ & $0.371\pm0.016$ & $0.309\pm0.080$ \\
  $\text{DRGPM}^*$ & $\cdots$ & $0.855\pm0.007$ & $\textbf{0.525}\pm0.022$ & $0.226\pm0.001$ & $0.284\pm0.017$\\ 
  CN & $0.189\pm0.035$ & $0.426\pm0.006$ & $0.121\pm0.009$ & $0.333\pm0.065$ & $0.138\pm0.017$ \\
  $\text{SVD++}^*$ & $\cdots$ & $0.423\pm0.026$ & $0.135\pm0.008$ & $0.214\pm0.016$ & $\cdots$ \\
  MNE & $0.315\pm0.018$ & $0.269\pm0.004$ & $0.227\pm0.014$ & $0.262\pm0.009$ & $0.347\pm0.037$ \\ 
  DeepWalk & $0.167\pm0.068$ & $0.191\pm0.009$ & $0.117\pm0.015$ & $0.252\pm0.019$ & $0.192\pm0.054$ \\ 
  \midrule
  Recurrent-DBN,K=30 & $\textbf{0.569}\pm0.022$ & $\textbf{0.881}\pm0.003$ & $0.509\pm0.017$ & $\textbf{0.543}\pm0.022$ & $\textbf{0.373}\pm0.016$ \\
  Recurrent-DBN,K=20 & $\textbf{0.476}\pm0.081$ & $\textbf{0.869}\pm0.003$ & $0.468\pm0.013$ & $\textbf{0.469}\pm0.026$ & $\textbf{0.361}\pm0.016$ \\ 
  %\textbf{Recurrent-DBN,KK=15} & $0.549\pm0.079$ & $\cdots$ & $0.867\pm0.004$ & $0.453\pm0.016$ & $0.423\pm0.040$ \\
  Recurrent-DBN,K=10 & $\textbf{0.457}\pm0.042$ & $0.853\pm0.007$ & $0.450\pm0.014$ & $0.369\pm0.010$ & $\textbf{0.356}\pm0.016$ \\
  \bottomrule
\end{tabular}
\label{auc_dataset}
\end{table*}

\section{Related Work}
Several Bayesian deep probabilistic frameworks have been proposed to capture the temporal dependence in dynamic data~\cite{gan2015deep,gong2017deep,henao2015deep}. The Deep Dynamic Sigmoid Belief Network~\cite{gan2015deep} sequentially stacks models of sigmoid belief networks and uses the binary-valued hidden variables to depict the log-range dynamic dependence. The Deep Dynamic Poisson Factor Analysis (DDPFA)~\cite{gong2017deep} incorporates the Recurrent Neural Networks (RNN) into the Poisson Factor Analysis~(PFA) to depict the long-range dynamic dependence. However, in DDPFA, the parameters in RNN and the latent variables in PFA are optimized separately. Poisson Gamma Dynamic Systems (PGDS) \cite{schein2016poisson} are developed to model the counting data through a ``shallow'' modelling strategy. Dynamic-PGDS~(DPGDS)~\cite{guo2018deep} is probably the closest work to our approach. Compared with DPGDS, our Recurrent-DBN differs in three aspects: (1) our Recurrent-DBN generates normalized latent representations and thus provides more interpretable structures; (2) the count information is propagated in a different way; (3) our Recurrent-DBN is devised in the setting of relational modelling, while DPGDS is for the topic modelling setting. 

For modelling dynamic network data, many of the existing works are ``shallow'' probabilistic modelling. The dynamic Tensorial Mixed Membership Stochastic Block model~(T-MBM)~\cite{tarres2019tensorial} and the Fragmentation Coagulation Based MMSB~(fcMMSB)~\cite{fcmmsb} combine the notable mixed-membership stochastic block model with a dynamic setting. The Bayesian Poisson Tensor Factorization~(BPTF)~\cite{schein2015bayesian} and the Dependent Relational Gamma Process model~(DRGPM)~\cite{yang2018dependent} are the representative works that use Poisson matrix factorization techniques to address dynamic counting data. There are also some models using the collaborative filtering techniques such as SVD++. Some methods are not developed for dynamic network data originally, but they have later been applied to the dynamic scenario, such as structure-based models like Common Neighbor~(CN)~\cite{newman2001clustering}, and network embedding models, including Scalable Multiplex Network Embedding~(MNE)~\cite{zhang2018scalable} and DeepWalk~\cite{perozzi2014deepwalk}. It is noted that there is a recent trend in using the graphon theory~\cite{RandomFunPriorsExchArrays,orbanz2014bayesian} to model the network data~\cite{xuhui2016OstomachionProcess,pmlr-v84-fan18b,NIPS2018_RBP,pmlr-v89-fan18a,online_BSPF}.

%\textbf{Computational complexity} {\color{red}I will add the computational complexity.}
%The probilistic models achieve higher accuracy although these methods require more computation time to collect MCMC samples. ... is much faster than the probabilistc models because its inference is performed using .... Table xx compares the per-iteration computation time of the sampling-based models (all models are implemented in ...). The computational cost of ... scales in $\mathcal{O}()$. The Bernoulli-Poisson link based models (.., DPGM, DRGPM) are much faster than the logistic link based method (DRIFT) because the former models scale linearly with the number of non-zero entries in network data. For Recurrent-DBN, sampling ... and ... takes $\mathcal{O}()$ with Ne being the number of non-zero entries. Sampling $\{C_(i,k,t)\}_{i,k,t}$ takes $\mathcal{O}()$ and sampling
%$\{\lambda_{k_1k_2}\}_{k_1,k_2}$ takes $\mathcal{O}()$. Overall, the computational complexity of Recurrent-DBN is $\mathcal{O}()+ \mathcal{O}()$. The computational complexity of D-GPPF and DPGM is $\mathcal{O}()$ and $\mathcal{O}()$, respectively. ... is slightly faster than ... because ... can effectively ... and hence achieves a lower computational cost.
%Among various node-neighbor-based indices, Common Neighbors (CN) is undoubtedly the precursor with low computing complexity. It has also been revealed that CN achieves high prediction accuracy compared with other classical prediction indices [25]. CN, however, only emphasizes the number of common neighbors but ignores the difference in their contributions.

\section{Experiments}
We evaluate the performance of our proposed Recurrent-DBN on five real-world data sets, by comparing with nine baseline methods: Mixed Membership Stochastic Block model~(MMSB)~\cite{airoldi2008mixed}, T-MBM, fcMMSB, BPTF, DRGPM, SVD++, CN, MNE and DeepWalk.
%and the Fragmentation Coagulation Based MMSB~(fcMMSB), involving the Mixed Membership Stochastic Block model~(MMSB)~\cite{airoldi2008mixed}; two Poisson matrix factorization model, containing the Bayesian Poisson Tensor Factorization~(BPTF)~\cite{schein2015bayesian} and the Dependent Relational Gamma Process model~(DRGPM)~\cite{yang2018dependent}; one approach for dynamic data used in recommendation systems, which is the collaborative filtering with temporal dynamics~(SVD++)~\cite{koren2009collaborative}; one structure-based models as Common Neighbor~(CN)~\cite{newman2001clustering}; and two network embedding models, including the Scalable Multiplex Network Embedding~(MNE)~\cite{zhang2018scalable} and the DeepWalk~\cite{perozzi2014deepwalk}. 
Except MMSB, all of the other eight baseline models are implemented with the released code. For MMSB, we use Gibbs sampling for the inference of all variables.

\subsection{Data set and experimental setting}
The real-world relational data sets used in this paper are: Coleman~\cite{coleman1964introduction}, %records the closest friends for each student at different years and seasons; 
Mining Reality~\cite{eagle2006reality}, %records contact data of selected students at the Massachusetts Institute of Technology over ten continuous months in the same year, where the link is set to $1$ if two student have at least one contact at that time step;
Hypertext~\cite{isella2011s}, % records the contact network of the ACM Hypertext 2009 conference attendees, and sets the relation as $1$ when two attendees have a face-to-face contact over 20 seconds;
Infectious~\cite{isella2011s} % describes the face-to-face interactions between people during the exhibition, and also the relation was recorded as $1$ as described as data set Infectious. 
and Student Net~\cite{fan2014dynamic}. %describes the relations between students among eleven time steps,
The summarized statistics are detailed in Table~\ref{rm_dataset}. For the hyper-parameters, we specify $M \sim \text{Gamma}(N, 1)$ for all data sets, $\{c_{c}^{(l)},c_{u}^{(l)}\}_{l}, d, d_c$ and $\Lambda_{k1,k2}$ are all given $\text{Gamma}(1, 1)$ priors and $L=3$. %We randomly extract a subset of $10\%$ network entries (either links or non-links) from each data set at each time step and set as test. The remaining $80\%$ in construct as the relational data matrix for training to predict the test entries. The test relational data are not used when constructing the information propagation matrix (i.e. we set $\{\beta_{i'i,t}^{(l)}\}_{l,t} = 0$ if $R_{i'i}$ is testing data).
For MMSB, we set the membership distribution %$\pmb{\theta}$ 
according to $\text{Dirichlet}(\pmb{1}^{1\times K})$.

\begin{table}[t]
\small
\centering 
\begin{tabular}{l r r r r}
  \toprule
  Data set & $N$ & $T$ & $N_E$ & $S\%$ \\
  \midrule
  Coleman & $73$ & $2$ & $506$ & $4.75$ \\ 
  Mining reality & $96$ & $10$ & $15580$ & $16.9$ \\
  Hypertext & $113$ & $10$ & $6996$ & $5.48$ \\
  Infectious & $410$ & $10$ & $7112$ & $0.42$ \\
  Student net & $1005$ & $11$ & $62041$ & $0.56$ \\
  \bottomrule
\end{tabular}
\caption{Data set information. $N$ is the number of nodes, $T$ is the number of time steps, $N_E$ is the number of positive links and $S\%$ is the ratio of the number of positive links to the total number of links.}
\label{rm_dataset}
\end{table}
%\subsection{Experimental setting} 

\subsection{Link prediction}

For link prediction, we randomly extract a proportion of $10\%$ of relational data entries (either links or non-links) at each time step as the test set. The remaining $90\%$ is used for training. The test relational data are not used to construct the information propagation matrix (i.e., we set $\{\beta_{i'i,t}^{(l)}\}_{l,t} = 0$ if $R_{i'i}$ is the testing data).
We estimate the posterior mean of $e^{-{\sum_{k_1,k_2}X_{i,k_1,t}\Lambda_{k_1k_2}X_{j,k_2,t}}}$ as the linkage probability for each test data. These linkage probabilities are then used to calculate two evaluation metrics: the area under the curve of the receiver operating characteristic~(AUC) and the precision-recall~(precision). Higher values of AUC and precision indicate better model performance.

The detail results are shown in Table~\ref{auc_dataset}. We report the average evaluation results for each model over $16$ runs. Each run uses $3000$ MCMC iterations with the first $1500$ discarded as burn-in. Overall, Recurrent-DBN outperforms the baseline models for both metrics on almost all data sets. As might be expected, the value of AUC and precision increase with higher model
complexity of Recurrent-DBN (i.e., larger values of $K$). For the other methods, fcMMSB is competitive with DRGPM and outperforms the other baselines. However, they all perform worse than Recurrent-DBN, especially for data sets with large numbers of $N$ or $T$. We can see that the Recurrent-DBN has clear advantages in learning dynamic relational data, thanks to the deep hierarchical structure and recurrent long-term temporal dependence modelling.

\subsection{Latent variable visualization}
\begin{figure}[t]
\centering
\includegraphics[width = 0.43\textwidth]{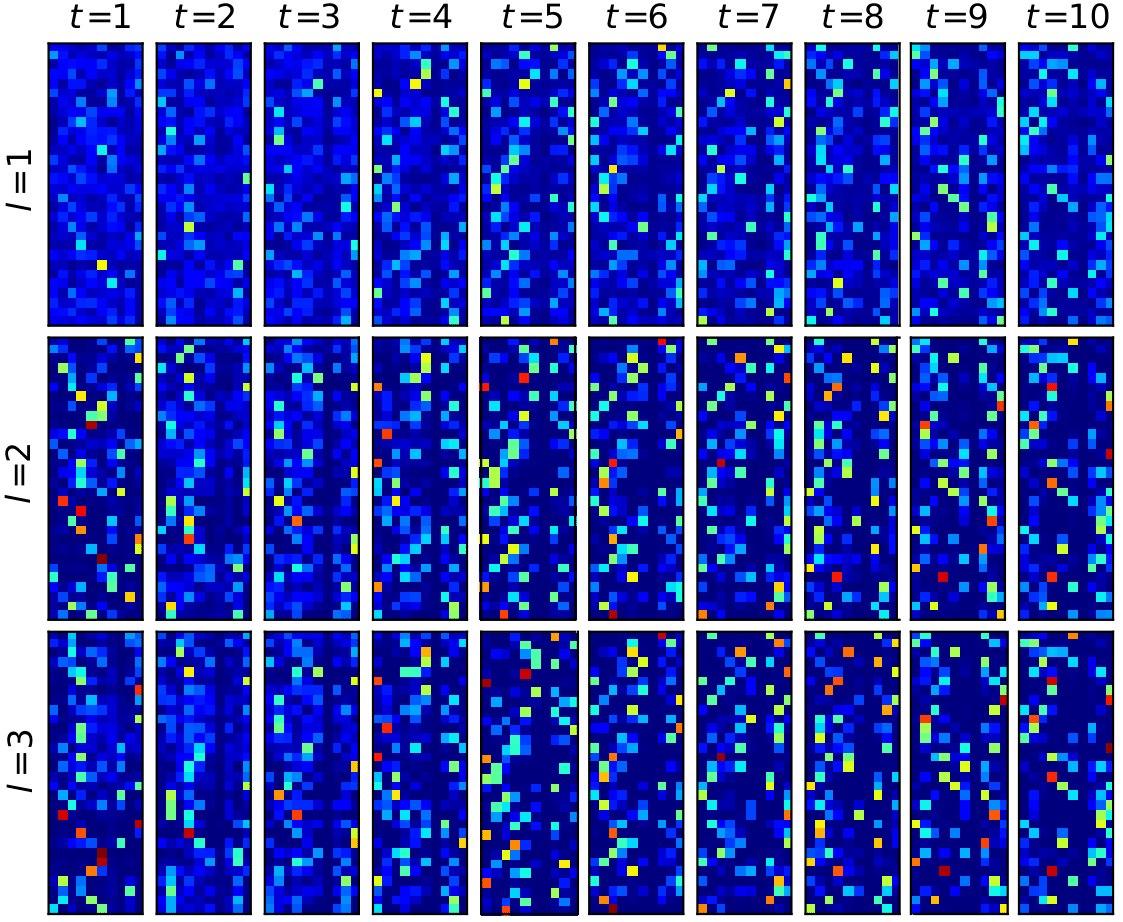}
\includegraphics[width = 0.4\textwidth]{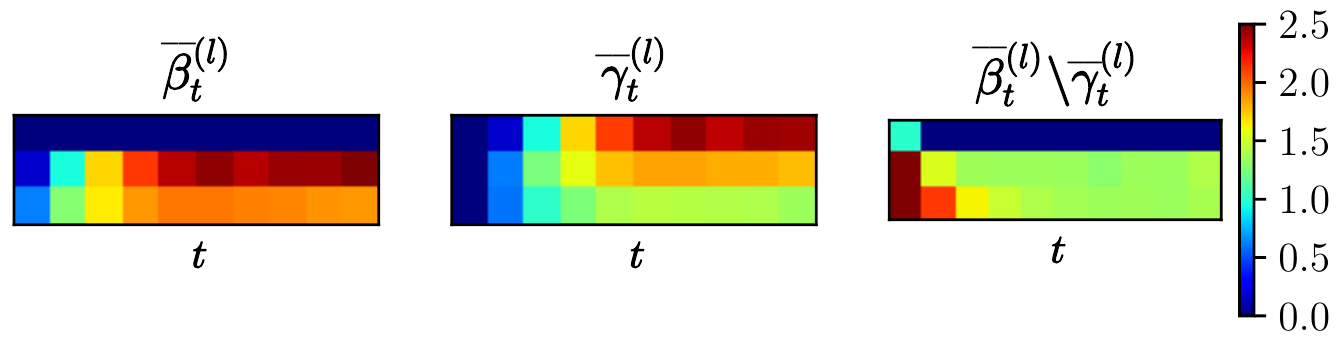}
\caption{Top: visualizations of the membership distributions ($\{\pmb{\pi}_{i=1:30}^{(l)}\}_{l=1}^{3}$) for the {\it Infectious} data set. Rows represent the nodes and columns represent the communities~(with $K=10$); Bottom: visualizations of average propagation coefficients $\overline{\pmb{\beta}}_{t}^{(l)},\overline{\pmb{\gamma}}_{t}^{(l)}$ and their ratio. $\overline{\pmb{\beta}}_{t}^{(l)},\overline{\pmb{\gamma}}_{t}^{(l)}$ are re-scaled for visualization convenience.}
\label{fig:pi_visulization}
\end{figure}

To gain further insights, we visualize the latent variables in Figure~\ref{fig:pi_visulization}. It can be observed from the top part that: (1) for the same time step, the membership distributions change gradually with the increase of layers; (2) the membership distributions share some similarities for consecutive time steps and the similarities slowly shift along with the time. For example, the left bottom area of $\{\pmb{\pi}_{i=1:30}^{(3)}\}$ seems to have $3$ different patterns: time steps $t=1$, $t=2\sim4$, and $t=5\sim 10$. The bottom part of Figure~\ref{fig:pi_visulization} visualizes propagation coefficients. It is reasonable to see the values of $\overline{\pmb{\beta}}$ in the first layer and $\overline{\pmb{\gamma}}$ in the first several time steps are small, since less information is propagated in these cases. The values become larger when more information is propagated. Also, the layer-wise propagation seems to have a larger influence than the cross-time propagation, with an average value of $\overline{\pmb{\beta}}/\overline{\pmb{\gamma}}=1.2\sim1.4$.

\section{Conclusion}
We have presented a probabilistic deep hierarchical structure named Recurrent Dirichlet Belief Networks~(Recurrent-DBN) for learning dynamic relational data. Through Recurrent-DBN, the evolution of the latent structure is characterized by both the cross-layer and the cross-time dependencies. We also develop an upward-backward–forward-downward information propagation to enable  efficient Gibbs sampling for all  variables. 
The experimental results on a variety of real data sets demonstrate the excellent predictive performance of our model, and the inferred latent structure provides a rich interpretation for both hierarchical and dynamic information propagation. Our Recurrent-DBN can be applied to tasks like dynamic topic models~\cite{guo2018deep,zhao2018dirichlet}) and dynamic collaborative filtering. We keep these potential applications as the future work.

\section*{Acknowledgements}
This work is partly supported by ARC Discovery Project DP180100966. Yaqiong Li is a recipient of UTS Research Excellence Scholarship. Xuhui Fan and Scott A.~Sisson are supported by the Australian Research Council through the Australian Centre of Excellence in Mathematical and Statistical Frontiers (ACEMS, CE140100049). Bin Li is supported by Shanghai Municipal Science \& Technology Commission (16JC1420401) and the Program for Professor of Special Appointment (Eastern Scholar) at Shanghai Institutions of Higher Learning.

\bibliographystyle{named}
\bibliography{ijcai20}

\end{document}